\documentclass[letterpaper, 10 pt, conference]{ieeeconf}  

\IEEEoverridecommandlockouts                              

\usepackage{times}
\usepackage{epsfig}

\usepackage{color} 
\usepackage{amsmath}
\usepackage{amssymb}
\usepackage{graphicx}
\usepackage{multirow}
\usepackage{tabularx}
\usepackage{adjustbox}
\usepackage[caption=false,font=footnotesize]{subfig}
\usepackage{siunitx}
\usepackage{booktabs}

\usepackage{stmaryrd} 
\usepackage{latexsym} 

\usepackage{url}
\usepackage{hyperref} \hypersetup{colorlinks=true, linkcolor=black, urlcolor=RubineRed}
\usepackage{kotex} 

\newcommand{\figref}[1]{Fig.~\ref{#1}}

\newcommand{\tabref}[1]{Tab.~\ref{#1}}

\newcommand{\ie}{\textit{i.e.}}
\newcommand{\eg}{\textit{e.g.}}

\usepackage[dvipsnames]{xcolor}

\title{\Large \bf DRL-ISP: Multi-Objective Camera ISP with Deep Reinforcement Learning}

\author{Ukcheol Shin*, Kyunghyun Lee*, and In So Kweon %
\thanks{*Both authors contributed equally to this work.}
\thanks{This work was conducted by Center for Applied Research in Artificial Intelligence (CARAI) grant funded by DAPA and ADD (UD190031RD).}
\thanks{U. Shin, K. Lee, and I. S. Kweon are with the School of Electrical Engineering, KAIST, Daejeon 34141, Republic of Korea. 
{\tt\small \{shinwc159, kyunghyun.lee, iskweon77\}@kaist.ac.kr}}%
}

\begin{document}

\maketitle
\thispagestyle{empty}
\pagestyle{empty}

\begin{abstract}
In this paper, we propose a multi-objective camera ISP framework that utilizes Deep Reinforcement Learning (DRL) and camera ISP toolbox that consist of network-based and conventional ISP tools. 
The proposed DRL-based camera ISP framework iteratively selects a proper tool from the toolbox and applies it to the image to maximize a given vision task-specific reward function. 
For this purpose, we implement total 51 ISP tools that include exposure correction, color-and-tone correction, white balance, sharpening, denoising, and the others.
We also propose an efficient DRL network architecture that can extract the various aspects of an image and make a rigid mapping relationship between images and a large number of actions.
Our proposed DRL-based ISP framework effectively improves the image quality according to each vision task such as RAW-to-RGB image restoration, 2D object detection, and monocular depth estimation. 
\end{abstract}


\section{Introduction}

In recent years, the importance of vision sensors has been re-emphasized as deep learning has demonstrated superior performance in various computer vision tasks.
Despite the importance, visible-light cameras suffer from hardware limitations such as narrow dynamic range and low sensor sensitivity.
For this problem, the conventional camera performs a built-in Image Signal Processing (ISP) that improves image quality by applying sequential modifications such as de-blurring, de-noising, and color enhancement.
However, the built-in ISP usually consists of a fixed image processing pipeline with factory-tuned hyperparameters. 
Therefore, the built-in ISP usually does not guarantee an optimal quality image for various computer vision tasks. 

On the other hands, recent deep-learning based approaches shows notable results such as direct RAW-to-RGB recovery~\cite{ignatov2020replacing,schwartz2018deepisp}, denoising~\cite{zhang2017beyond}, super resolution~\cite{kim2016accurate, tong2017image}, white-balance~\cite{afifi2019color,afifi2020deep}, tone-mapping~\cite{liu2020single,marnerides2018expandnet}, and exposure correction~\cite{afifi2020learning,hu2018exposure,yu2018deepexposure} through a single deep neural network.
However, they require high computational costs and can only replace a specific part of the camera ISP pipeline.

Based on the observation, we propose a new camera ISP framework that utilizes Deep Reinforcement Learning (DRL) and camera ISP toolbox that includes both traditional image processing tools and network-based tools.
Our DRL framework applies the most appropriate ISP tools based on the current image state to maximize a target reward function with a given demosaiced RAW image.
Based on designed reward functions, the DRL agent can generate an image suitable for various tasks, such as general RAW-to-RGB recovery, object detection, and depth estimation.

Our contributions include the following:
\begin{itemize}
    \item We propose a novel DRL-based camera ISP framework that effectively performs a suitable action according to the current image state and target reward function.
    \item We propose a camera ISP toolbox along with its training method. 
    The toolbox consists of light-weight CNN tools and traditional tools that can represent each block of the camera ISP pipeline.
    \item We propose an efficient DRL network architecture that extracts various aspects of an image and build rigid mapping relationships between images and a large number of action space.
    \item We validate our proposed method for RAW-to-RGB image restoration, 2D object detection, and monocular depth estimation tasks. 
    The proposed method consecutively increases the performance of the target task by modifying images suitably.
\end{itemize}

\begin{figure}[!t]
\begin{center}
\footnotesize
\begin{tabular}{c@{\hskip 0.01\linewidth}c@{\hskip 0.01\linewidth}c}
\multicolumn{3}{c}{\includegraphics[width=0.96\linewidth]{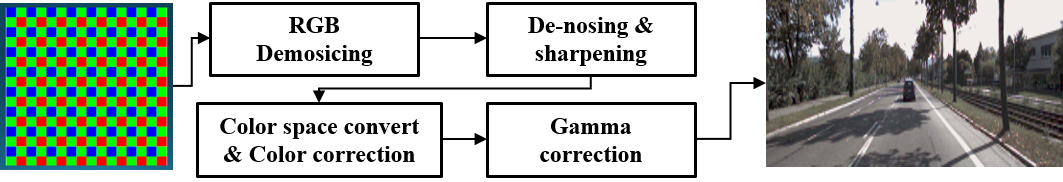}} \\ 
\multicolumn{3}{c}{\bf (a) Traditional Camera ISP Pipeline~\cite{abdelhamed2018high}} \\ 
\multicolumn{3}{c}{\includegraphics[width=0.96\linewidth]{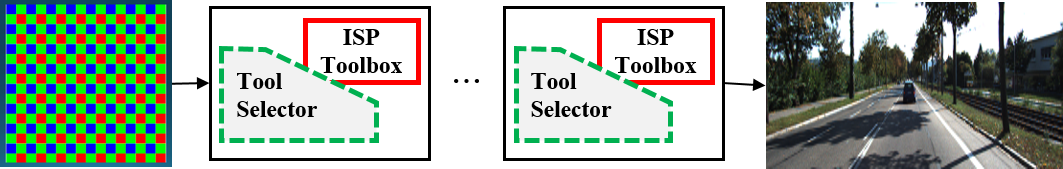}} \\ 
\multicolumn{3}{c}{\bf (b) Proposed DRL based Deep Camera ISP Pipeline} \\ 
\includegraphics[width=0.32\linewidth]{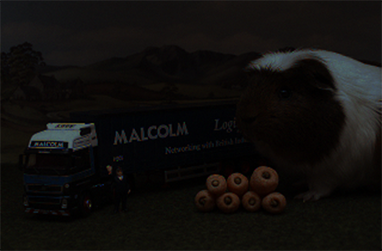} & 
\includegraphics[width=0.32\linewidth]{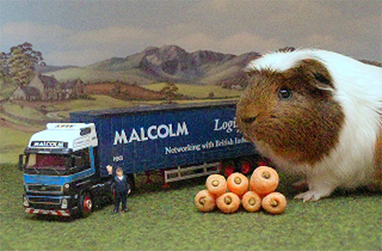} & 
\includegraphics[width=0.32\linewidth]{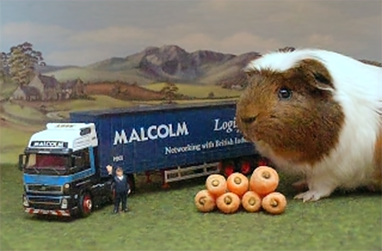} \\ 
{\footnotesize $1^{st}$: Input} & {\footnotesize $2^{nd}$: Color, Tone} & {\footnotesize $3^{rd}$: Denoise} \\
\includegraphics[width=0.32\linewidth]{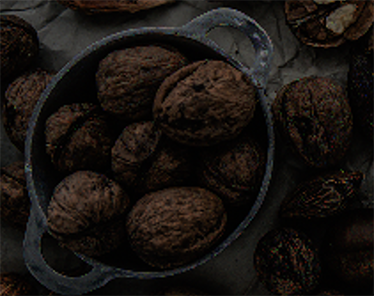} & 
\includegraphics[width=0.32\linewidth]{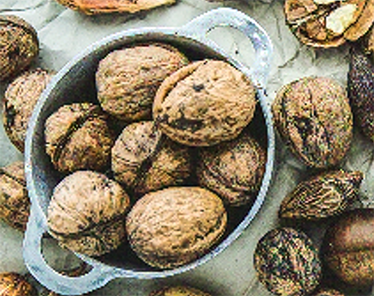} & 
\includegraphics[width=0.32\linewidth]{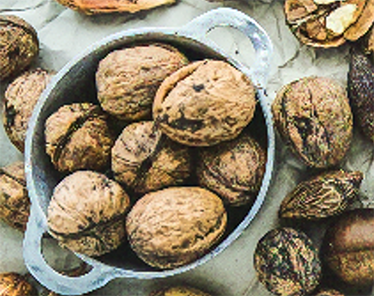} \\ 
{\footnotesize $1^{st}$: Input} & {\footnotesize $2^{nd}$: Color, Tone} & {\footnotesize $3^{rd}$: DeJPEG} \\
\multicolumn{3}{c}{\bf (c) RAW-to-RGB with PSNR reward} \\ 
\end{tabular}
\end{center}
\caption{{\bf Overview of the proposed DRL-based Camera ISP framework}. 
The traditional camera ISP framework (a) usually consists of a non-flexible image processing pipeline with factory-tuned hyperparameters.
On the other hand, the proposed DRL-based ISP framework (b) has a flexible pipeline that can adaptively process a given image by selecting desirable ISP tools sequentially (c). 
}
\label{fig:teaser}
\end{figure}

\section{Related Work}
\subsection{Camera ISP Parameter Optimization}
Traditionally, the RGB image is recovered from the RAW image through a camera built-in ISP chipset that consists of various image processing blocks.
Recently, several approaches have explored the automatic camera ISP optimization with various objectives~\cite{nishimura2018automatic,tseng2019hyperparameter,yahiaoui2019optimization,dong2018learning,hevia2020optimization,mosleh2020hardware,yang2020effective}.
Each image processing block of the ISP chipset is usually a black-box.
Therefore, some researches optimize chipset's hyperparameters through black-box optimization~\cite{nishimura2018automatic}, evolutionary algorithm~\cite{hevia2020optimization}, and reinforcement learning~\cite{mosleh2020hardware}.
Other approaches~\cite{dong2018learning, tseng2019hyperparameter} parameterize the operation of each block or entire ISP pipeline as a neural network. 
After that, they optimize the hyperparameter through the approximated neural network. 

Sometimes, they optimize hyperparameters for low-level image enhancement~\cite{dong2018learning,tseng2019hyperparameter}, object detection~\cite{yahiaoui2019optimization}, human preference~\cite{yang2020effective}, and high-level scene understanding~\cite{buckler2017reconfiguring,mosleh2020hardware}.
However, Their methods rely on a fixed ISP pipeline and hyperparameters.
Therefore, they cannot easily add a new image processing module and change the parameters according to a new image or new environment adaptively.



\subsection{Learning Camera ISP Pipeline}
Recently, neural network based approaches~\cite{chen2018learning,ignatov2020replacing,schwartz2018deepisp,liang2021cameranet,dai2020awnet} are emerging to directly recover a high-quality RGB image from RAW sensory data through a single deep neural network. 
Their underlying idea is to embed the entire ISP pipeline, including demosaicing, denoising, sharpening, color correction, and white balance, into a single deep neural network.
Mainly, their objective is well-exposed images~\cite{schwartz2018deepisp, chen2018learning}, expert retouched images~\cite{liang2021cameranet}, and well-captured camera images~\cite{ignatov2020replacing,dai2020awnet}.

Contrarily, some studies have been proposed to approximate only a specific part of the ISP pipeline, such as white-balance~\cite{afifi2019color,afifi2020deep}, tone-mapping~\cite{liu2020single,marnerides2018expandnet}, and exposure correction~\cite{afifi2020learning,hu2018exposure,yu2018deepexposure}.  
These entire and partial replacements show higher performance than typical ISP pipelines and specific modules.
However, the networks are usually computationally heavy and impossible to specialize for non-differentiable objectives (\eg well-exposed and edge-preserving image, object detection, or high-level scene understanding).
\begin{figure*}[t]
\begin{center}
\footnotesize
\begin{tabular}{c}
\includegraphics[width=0.95\linewidth]{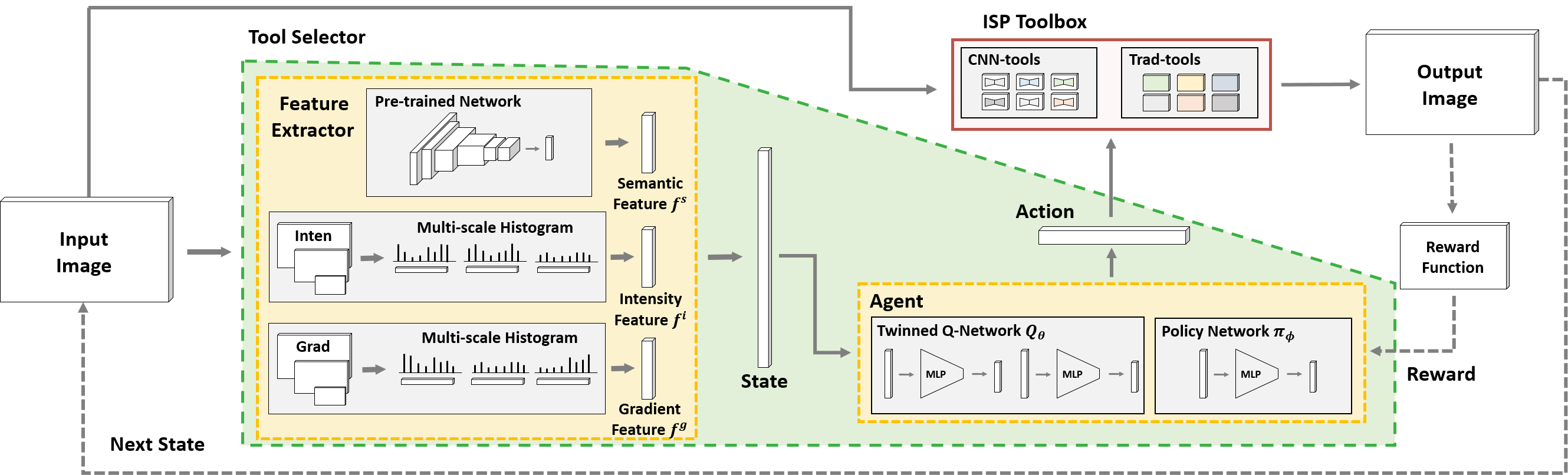} \\ 
\end{tabular}
\end{center}
\caption{{\bf Overall pipeline of the proposed DRP-ISP framework}. The proposed DRL-ISP framework  consists of a feature extractor $F$ and camera ISP toolbox. 
The ISP toolbox exploits both CNN-based and traditional image processing tools that can represent each block of the camera ISP pipeline.
With a given image input, the feature extractor $F$ extracts an efficient feature vector $f^{ag}$ that includes intensity, gradient, and semantic information.
The policy network $\pi_{\phi}$ selects optimal action according to the feature vector $f^{ag}$ to maximize a target reward function.
The twinned Q-networks$Q_{\theta}$ are only utilized during the training. 
}
\label{fig:overall_pipeline}
\vspace{-0.10in}
\end{figure*}

\section{Method Overview}
\subsection{Problem Definition}
We treat the camera ISP pipeline as a {\it sequential decision making problem} that iteratively decides proper action $a_t$ according to the current image state $s_t$.
Given a demosaiced RAW image, our goal is to make an enhanced optimal image $I_{opt}$ for a target reward function $R(\cdot)$.
We define each Image Signal Processing (ISP) tool as an action $a_t$, image feature $f^{ag}$ extracted from the demosaiced RAW image as a state $s_t$, and target-task specific objective function as a reward function $R(\cdot)$.

\subsection{Camera ISP Toolbox}
The proposed framework consists of camera ISP toolbox and tool selector, as shown in~\figref{fig:overall_pipeline} and~\tabref{tab:tool_box}.
We design the toolbox to include the functionality of each camera ISP module, such as white balance, denoising, sharpening, and color-and-tone correction.
The proposed toolbox exploits traditional and learning-based methods to leverages both high performances of the deep network and controllability of the traditional tools.
All network based tools are trained in a self-supervised manner. 

\subsection{DRL-based ISP Tool Selector}
The tool selector consists of a feature extractor and Deep Reinforcement Learning (DRL) agent. 
The feature extractor extracts representative feature vector $f^{ag}$ that consists of intensity, gradient, and semantic information.
After that, the DRL agent chooses a proper ISP tool to maximize the target reward function.
The DRL agent is trained with the policy gradient algorithm, proposed in~\cite{haarnoja2018soft}.

\begin{table}[t]
\caption{\textbf{Camera ISP toolbox specification}.
The ISP toolbox consists of both traditional and network-based image processing tools.
The number of $(\cdot)$ indicates each tool has the corresponding number of actions.
}
\begin{center}
\resizebox{0.98 \linewidth}{!}{
\def\arraystretch{1.2}
\footnotesize
\begin{tabular}{c|c|c}
\hline
Type & {\textbf{Network-based Tools}} & {\textbf{Traditional Tools}}    \\ 
\hline \hline
Brightness & ExposureNet (x2) & Brightness modification (x12)\\
\hline 
\multirow{3}{*}{Contrast} & \multirow{3}{*}{CTCNet (x2)} & Histogram equalization, \\
                                  &                          & CLAHE~\cite{reza2004realization}, \\
                                  &                          & Gamma correction (x6) \\
\hline
\multirow{3}{*}{Color} & \multirow{3}{*}{WBNet (x2)} & Hue modification (x6),\\
                               &                    & Saturation modification (x6),\\ 
                               &                    & White Balance~\cite{ebner2007color} \\ 
\hline
Noise & DenoiseNet (x2) & Gaussian, box, bilateral filter \\
\hline
Blur & DeblurNet (x2) & Sharpening filter \\
\hline
\multirow{2}{*}{Others} & SRNet (x2) & \multirow{2}{*}{Do nothing} \\ 
 & DejpgNet (x2) & \\ 

\hline
\end{tabular}
}
\end{center}
\label{tab:tool_box}
\end{table}

\begin{figure}[t]
\begin{center}
    \footnotesize
    \begin{tabular}{c@{\hskip 0.002\linewidth}c}
     \includegraphics[width=0.48\linewidth]{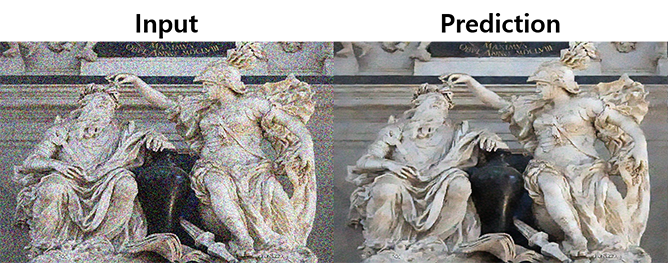} & 
     \includegraphics[width=0.48\linewidth]{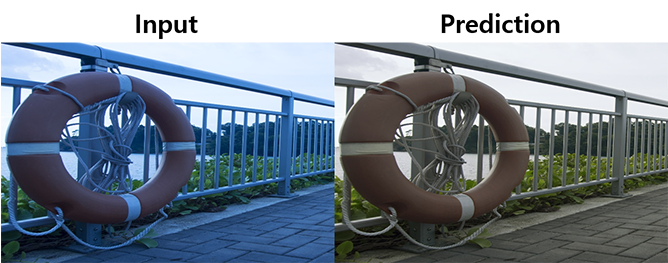} \\ 
     {\footnotesize (a) Denoise}& {\footnotesize(b) White-Balance} \\
     \includegraphics[width=0.48\linewidth]{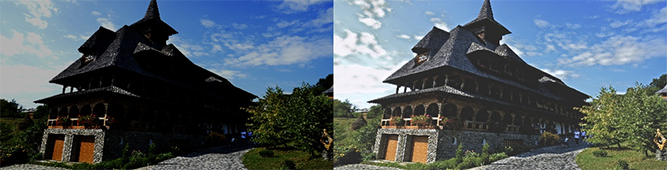} & 
     \includegraphics[width=0.48\linewidth]{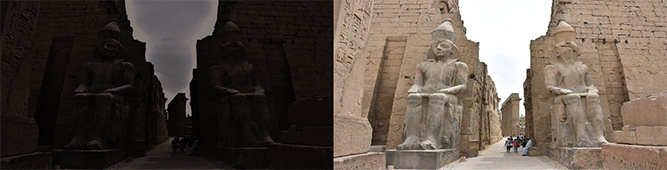} \\ 
     {\footnotesize (c) Exposure Correction}& {\footnotesize(d) Color-and-Tone Mapping} \\
     \includegraphics[width=0.48\linewidth]{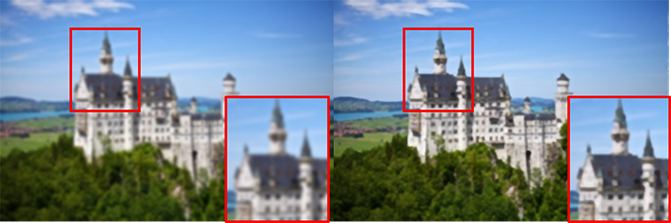} & 
     \includegraphics[width=0.48\linewidth]{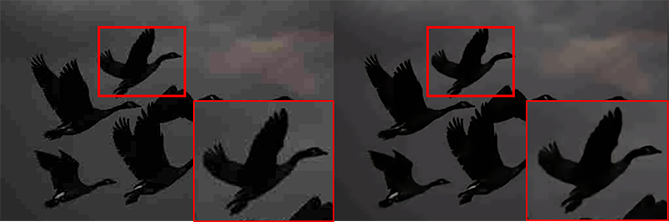} \\ 
     {\footnotesize (e) Deblur}& {\footnotesize(f) Dejpeg} \\
    \end{tabular}
\end{center}
\caption{{\bf Qualitative results of each CNN-based ISP tool.} 
Each CNN tool is trained to solve each target task, such as denoise, deblur, exposure correction, color mapping, and white balance, in a self-supervised manner.
}
\label{fig:cnntool}
\vspace{-0.10in}
\end{figure}

\section{Camera ISP Toolbox}
\label{sec:Method_toolbox}
The CNN-based tools show a high-representation capacity that encompasses complicated multi-step image processing pipeline while showing outperformed results.
However, they usually produce one-way mapping results, are non-controllable to get another result, and sometimes lead to undesirable results.
On the other hand, even though traditional tools do not perform better than the CNN tools, traditional image processing tools have been already demonstrated their computational efficiency, stability, and controllability.
Therefore, we design the camera ISP toolbox to include both traditional and learning-based tools for performance, stability, and controllability, as shown in~\tabref{tab:tool_box}.

\subsection{Traditional ISP Tools}
We implement brightness, contrast, color, noise, and blur handling methods from the openCV~\cite{bradski2008learning} and Korina library~\cite{eriba2019kornia}.
We consider this implementation can properly represents each block of camera ISP such as white-balance, denoising, sharpening, and color correction.
The same philosophy is incorporated into network-based tool designing, such as exposure correction, Color-and-Tone Correction(CTC), White-Balance(WB), denoise, deblur, Super-Resolution(SR), and de-jpeg network.

\subsection{Learning-based ISP Tools}
\subsubsection{Individual Tool Training.} 
In order to make light-weight network-based ISP tools, we adopt shallow 3-layer and 8-layer neural networks proposed in~\cite{yu2018crafting}.
We train each network in a self-supervised learning manner for each target task.
Given an original image $I_{ori}$, we make a distorted image $I_{dis}$ according to the target network tool type. 
For example, if we want to train the de-blur network, we apply blur kernel to the original image and make a distorted image.
After that, we train the networks to restore the original image $I_{res}$ from the distorted image $I_{dis}$ with the L1 loss $\ell_{L1}$ and feature reconstruction loss $\ell_{feat}$~\cite{johnson2016perceptual}, as follows:
\begin{equation} 
\label{equ:Individual tool trainig}
L_{ind}=\alpha\cdot\ell_{L1}(I_{res},I_{ori})+\beta\cdot\ell_{feat}(I_{res},I_{ori}),
\end{equation}
where $\alpha$ and $\beta$ are scale factors for each loss function.

\begin{figure}[t]
\begin{center}
\footnotesize
\begin{tabular}{c}
\includegraphics[width=0.9\linewidth]{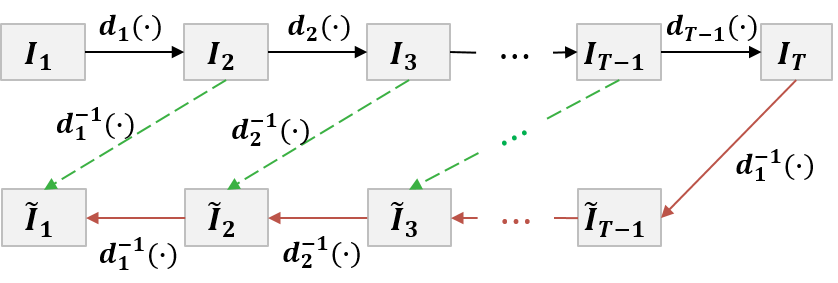} \\ 
\end{tabular}
\end{center}
\caption{{\bf Image distort-and-restore process for the collective learning of ISP tools}. $d_t(\cdot)$ indicates random image distortion function and $d_t^{-1}(\cdot)$ denotes corresponding image restoration tool. The red-line and green dash-line represents global and local restoration trajectories, respectively.}
\label{fig:joint_learning}
\end{figure}

\subsubsection{Collective Tool Training.} 
As described in~\cite{zhang2019deep,yu2018crafting}, the cascading of two separately trained networks usually cause undesirable unseen artifact. 
Therefore, we train all network-based tools collectively to relieve the artifact problem in a self-supervised manner.
As shown in~\figref{fig:joint_learning}, the underlying idea is to train the network-based tools to deal with the unseen artifacts without forgetting their pre-learned restore ability.
Compared to~\cite{yu2018crafting}, our collective tool training exploits both global and local restoration trajectories as follows:
\begin{equation} 
\label{equ:collective tool trainig}
L_{col} = \epsilon\cdot\ell_{L1}(\Tilde{I}_{1}^{G},I_{1})+(1-\epsilon)\cdot\Sigma_{t=1}^{T-1}\ell_{L1}(\Tilde{I}_{t}^{L},I_{t}),
\end{equation}
where $\epsilon$ is a scale factor, $\Tilde{I}_{1}^{G}$ indicates the recovered image following the global trajectory, and $\Tilde{I}_{1}^{L}$ indicates the recovered image following the local trajectory. 
Based on the collective tool training, the networks can handle the unseen artifacts while enhancing global restoration and preserving their pre-learned ability with local restoration trajectory.


\section{DRL based Tool Selector}
The proposed tool selector consists of a feature extractor $F$ and a DRL agent, as shown in~\figref{fig:overall_pipeline}.
At the beginning of each episode, a demosaiced RAW input image $I_{1}$ is fed to the tool selector.
In each timestep $t$, the feature extractor $F$ extracts the feature vector $f^{ag}_t=[f^{i}_t,f^{g}_t,f^{s}_t]$. 
The policy network $\pi_{\phi}$ of the DRL agent chooses a proper action $a_t\sim \pi(f^{ag}_t)$. 
The corresponding tool $a_t$ is applied to the given image.
After that, a processed image $I_{t+1}=a_t(I_t)$ and a reward $r_{t+1} = R(I_t, I_{t+1})$ are given. 
The episode ends when the timestep $t$ reaches the maximum episode timestep $T$, or the STOP action is chosen.

\subsection{Feature Extractor}
In terms of DRL-agent, a well-represented state vector is a key prerequisite to deciding appropriate action.
We empirically found that the implicit feature learning network, such as convolutional neural network (CNN), tends to extract less informative feature vector that leads to overall low performance because of spares supervision of reinforcement learning, as validated in \tabref{tab:abl}-(a).
Therefore, we design an explicit feature extraction module that extracts each feature vector from the intensity, gradient, and semantic space of the given image to represent various image properties.

Given input image $I$ is converted to intensity and gradient image via gray-scale conversion and Sobel operator.
After that, we apply the multi-scale histogram method~\cite{wang2020learning} to the intensity and gradient image to extract global and local information.
For the semantic feature, we utilize ImageNet pre-trained Alexnet~\cite{krizhevsky2012imagenet}. 
Then, the state vector is decided by aggregating all feature vectors; $s_t={f_{t}^{ag}}$, where $f_{t}^{ag}$ is a aggregated feature vector from the intensity $f^{i}$, gradient $f^{g}$, and semantic feature $f^{s}$.
 
\subsection{DRL Agent}
The network architecture of our DRL agent is similar to original Soft Actor-Critic (SAC)~\cite{haarnoja2018soft}, except that we use discrete action space~\cite{christodoulou2019soft}. 
The DRL agent consists of a twinned Q-network $Q_{\theta}$ and a policy network $\pi_{\phi}$, as shown in~\figref{fig:overall_pipeline}. 
All networks take the feature vector $f^{ag}$ extracted from the feature extractor $F$. 
The twinned Q-network estimates two Q-value $q_1$, $q_2$, and the policy network estimates probabilities for all actions. 
During the training stage, the smaller Q-value is selected as a Q-value $Q_\theta(s_t)$ for the training stability.
Entire networks are trained based on the policy gradient algorithm, as follows: 
\begin{align*}
    J_\pi(\phi) = & E_{s_t\sim D}\left[ \pi_\phi(s_t)[\kappa \log(\pi_\phi(s_t))-Q_\theta(s_t)]\right], \\
    J_Q(\theta) = & E_{(s_t, a_t)\sim D}[\frac{1}{2}(Q_\theta(s_t) - (r(s_t, a_t) + \\  
     & \gamma \pi(s_{t+1})[Q_\theta(s_{t+1}) - \kappa \log (\pi(s_{t+1}))]), ]
\end{align*}
where $\gamma$ is a discount factor, $D$ indicates replay buffer, and $\kappa$ denotes entropy scale factor. 




\subsection{Reward}
A reward function is freely definable according to various purpose without considering differentiability.
In this section, we define representative reward functions according to the target application, such as RAW-to-RGB restoration, object detection, and single-view depth estimation.
The basic form of reward function is as follows : 
\begin{equation*}
    R(I_{t}, I_{t-1}) = r_s\left[M(I_t) - M(I_{t-1})\right],
\end{equation*}
where $M(\cdot)$ is a metric function and $r_s$ is a scaling factor for each metric.
Here, $r_s$ can be a negative value for some metrics.
The reward function calculates the difference between the metric values of the previous and current images.
The metric functions for various tasks are shown in~\tabref{tab:metric_functions}.

\begin{table}[t]
\centering
\caption{\textbf{Metric functions for various target tasks.} 
Each metric function can be utilized solely and combined with another function for collaborative effect. 
}
\label{tab:metric_functions}
\resizebox{0.98\linewidth}{!}
{

    \def\arraystretch{1.4}
    \begin{tabular}{c|c|c}
    \toprule
    {\bf Task}                   & {\bf Metric} & {\bf Metric fucntion} $M(I_t)$\\
    \midrule
    \multirow{3}{*}{RAW-to-RGB} & PSNR         & $10 \log_{10}\left(\frac{1.0}{\sqrt{\sum(I_t-I_{gt})^2}} \right)$  \\
    
    \cline{2-3}
                                 & Color        & $\sum||I_t-{RGB}_{target}||$\\
    \cline{2-3}
                                 & Intensity        & $\sum||I_t-{Gray}_{target}||$\\
    \midrule
    \multirow{2}{*}{Detection}   & PR           &  $\sum_{k=1}^n {\left(w_p \text{Pr}(I_t, I_{gt}) + w_r \text{Re}(I_t, I_{gt})\right)}$\\
    \cline{2-3}
                                 & SOPR         &  $\sum_{k=1}^n {\text{SO}(k)\left(w_p \text{Pr}(I_t, I_{gt}) + w_r \text{Re}(I_t, I_{gt})\right)}$\\
    \midrule
    \multirow{2}{*}{Depth}       & RMSE         & $\sqrt{\sum(\text{Depth}(I_t) - \text{Depth}_{\text{gt}}(I_{gt}))^2}$  \\
    \cline{2-3}
                                 & $\delta_1$   & $\delta_1(I_t, I_{gt})$  \\
    \bottomrule
    \end{tabular}
}
\end{table}

\subsubsection{RAW-to-RGB Restoration}
The most straight-forward metric for the RAW-to-RGB restoration task is to measure the difference between the restored and the original RGB images.
Based on the Ground-Truth(GT) RGB images, the DRL agent can learn proper image modification method that mimics the camera ISP.
For this purpose, we use PSNR criteria, a widely used image quality metric. 
Furthermore, we design non-reference based reward metrics such as desirable color and intensity metrics.




\subsubsection{RAW-to-RGB Restoration for Vision Task}
One of the most important feature of our algorithm is that we could define the reward function for various purpose. 
Therefore, we also validate our proposed framework for object detection and depth estimation tasks.
We design each task-specific reward such as Precision-Recall based metric for object detection and RMSE based metric for depth estimation task. 

{\bf Object Detection.}
We use Mask R-CNN~\cite{he2017mask} with Resnet-50 as a reference detection model. 
Usually, the mAP metric is used for evaluating object detection performances.
However, since the mAP metric requires multiple images to be calculated, we defined Precision-Recall (PR) and Small Object Precision-Recall (SOPR) metric functions. 
Both functions are calculated by weighted summation of precision $\text{Pr}$ and recall $\text{Re}$ values with the given bounding boxes.
SOPR metric gives more weight to the small objects that have small bounding box areas below a certain threshold.
$w_p$ and $w_r$ are scaling factors, $\text{SO}(k)$ is a multiplication factor for the $k^{th}$ object which has $1$ or the fixed scalar $w_{\text{so}}$ according to object size. 

{\bf Depth Estimation.}
We utilize SC-SfMlearner \cite{bian2019unsupervised} as a reference depth model for single-view depth estimation task.
Since the depth evaluation result is directly related to the predicted depth map quality, we adopt the representative depth evaluation metrics RMSE and $\delta_1$ as our metric functions.
We believe another depth evaluation metric and unsupervised image reconstruction loss could be alternatively utilized for the reward metric function.
\begin{figure*}[t]
\footnotesize
    \centering
    \begin{tabular}{c@{\hskip 0.002\linewidth}c@{\hskip 0.002\linewidth}c@{\hskip 0.002\linewidth}c@{\hskip 0.002\linewidth}c}
     \includegraphics[width=0.19\linewidth]{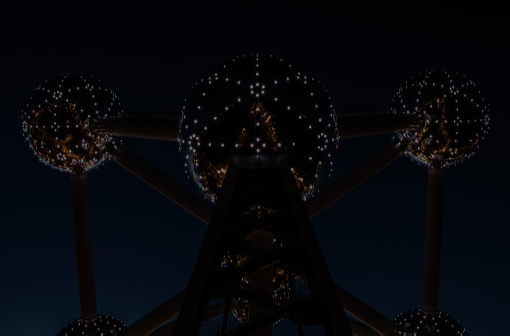} & 
     \includegraphics[width=0.19\linewidth]{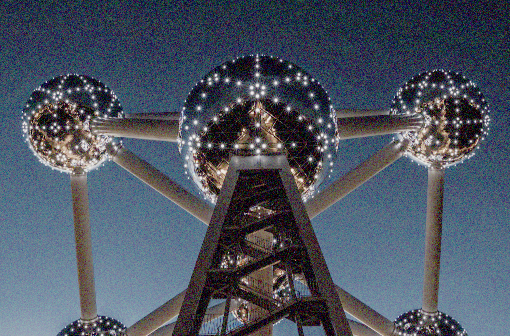} & 
     \includegraphics[width=0.19\linewidth]{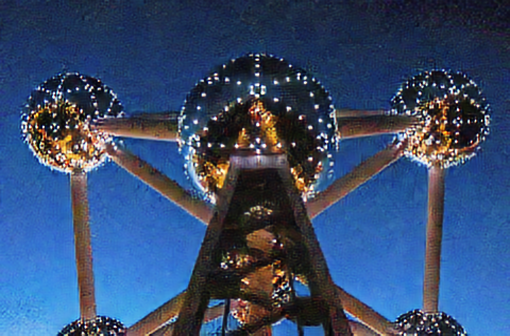} & 
     \includegraphics[width=0.19\linewidth]{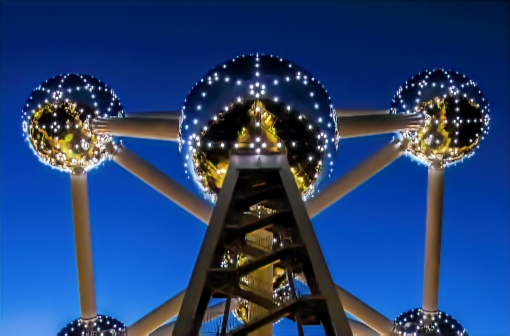} & 
     \includegraphics[width=0.19\linewidth]{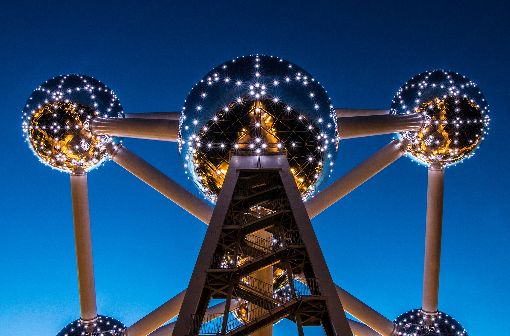} \\ 
     \includegraphics[width=0.19\linewidth]{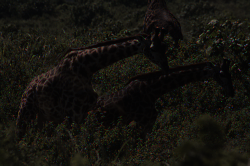} & 
     \includegraphics[width=0.19\linewidth]{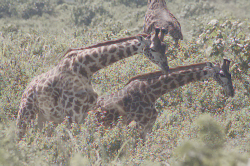} &
     \includegraphics[width=0.19\linewidth]{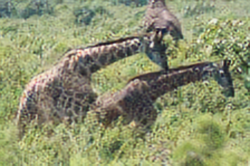} &
     \includegraphics[width=0.19\linewidth]{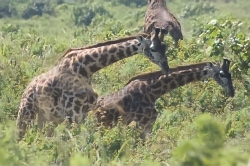} &
     \includegraphics[width=0.19\linewidth]{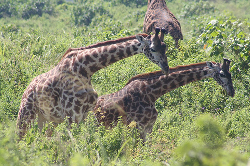} \\
     \includegraphics[width=0.19\linewidth]{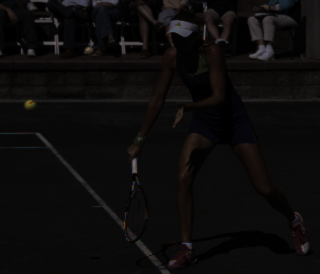} &
     \includegraphics[width=0.19\linewidth]{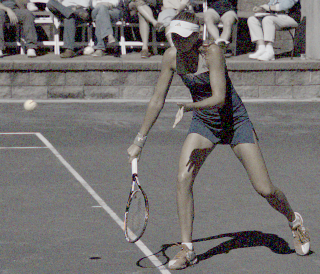} &
     \includegraphics[width=0.19\linewidth]{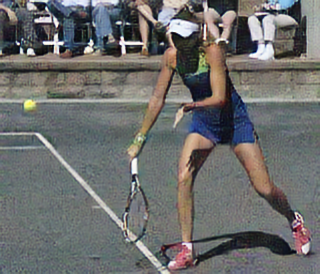} &
     \includegraphics[width=0.19\linewidth]{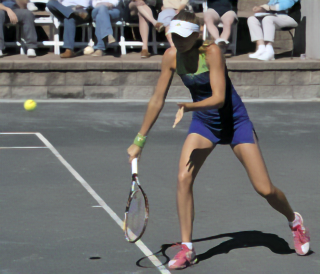} &
     \includegraphics[width=0.19\linewidth]{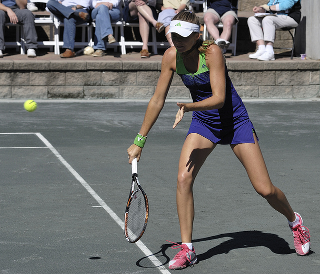} \\
      {\footnotesize (a) Input} & {\footnotesize (b) Default ISP~\cite{abdelhamed2018high}} & {\footnotesize (c) PyNet~\cite{ignatov2020replacing}} & {\footnotesize (d) Ours($M_{PSNR}$)} & {\footnotesize (h) GT label}\\
    \end{tabular}
\caption{{\bf Qualitative results for RAW-to-RGB processing task with various reward functions on synthetic datasets.} 
Ours proposed DRL-ISP framework effectively learns how to process an image according to the given reward function $M_{PSNR}$.
The proposed framework generates a clean, sharp, and colorful RGB image from the given RAW image by adaptively selecting proper ISP tools.
}

\label{fig:total RAW2RGB result}
\vspace{-0.10in}
\end{figure*}

\begin{table}[t]
\centering
\caption{\textbf{Quantitative results for RAW-to-RGB processing task.}
We train tool selectors according to PSNR reward function $M_{PSNR}$ on the synthetic RAW-to-RGB dataset. 
}
\vspace{-0.1in}
\resizebox{0.95\linewidth}{!}
{
    \def\arraystretch{1.1}
    \begin{tabular}{c|ccc}
    \hline
    \multirow{2}{*}{Methods} & \multicolumn{3}{c}{\textbf{Raw-to-RGB}} \\ 
    \cline{2-4}
     & PSNR & SSIM & MSSIM \\ 
    \hline \hline
     Default Camera ISP~\cite{abdelhamed2018high} & 19.88 & 0.6540 & 0.8853 \\
     PyNet~\cite{ignatov2020replacing}            & 23.24 & 0.7214 & 0.9246 \\
     Ours($M_{PSNR}$)                             & \textbf{24.82} & \textbf{0.7559} & \textbf{0.9375} \\
    \hline
    \end{tabular}
}
\label{tab:exp_raw2rgb}
\vspace{-0.15in}
\end{table}

\section{Experimental Results}
\subsection{Implementation Details}
{\bf Camera ISP Toolbox.}
In order to train CNN-based ISP tools, we construct a dataset that consists of MS-COCO~\cite{lin2014microsoft}, KITTI~\cite{geiger2012we}, and DIV2K~\cite{Agustsson_2017_CVPR_Workshops} dataset.
For all CNN-based ISP tools except the white-balance tool, we randomly selected 7,000 images from the training set and 1100 images from the testing set.
After that, the original images are distorted for the target ISP tool training by adding 
noise, blur, jpeg compression, resizing, brightness jittering effects, and inverting camera pipeline~\cite{buckler-iccv2017}.
The two types of networks (3-layer and 8-layer) are trained with low-level and high-level distortion, respectively.
For the white-balance dataset, we utilize the rendered WB dataset(Set2)~\cite{afifi2019color}.

{\bf ISP Tool Selector.}
We utilizes the above-mentioned ISP toolbox dataset for the DRL agent training.
However, the MS-COCO~\cite{lin2014microsoft}, KITTI~\cite{geiger2012we}, and DIV2K~\cite{Agustsson_2017_CVPR_Workshops} dataset doesn't provide RAW image. 
Therefore, firstly, we convert the original RGB image to a RAW Bayer image by utilizing the camera pipeline reversion method~\cite{buckler-iccv2017}.
After that, random augmentation such as brightness, noise, and blur effect is added to the converted RAW images to reflect a real-world capturing process.
Further details such as hyper-parameter are described in the supplementary video.

\begin{center}
    \begin{figure*}[tb]
    \footnotesize
        \begin{tabular}{c@{\hskip 0.002\linewidth}c@{\hskip 0.002\linewidth}c@{\hskip 0.002\linewidth}c@{\hskip 0.002\linewidth}c@{\hskip 0.002\linewidth}c@{\hskip 0.002\linewidth}c@{\hskip 0.002\linewidth}c}
        \multicolumn{2}{c}{\includegraphics[width=0.22\linewidth]{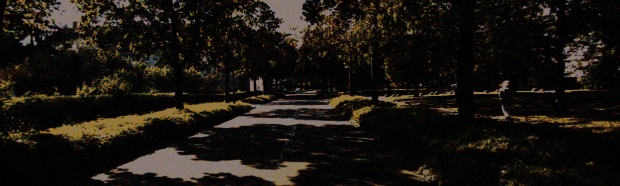}} & 
        \multicolumn{2}{c}{\includegraphics[width=0.22\linewidth]{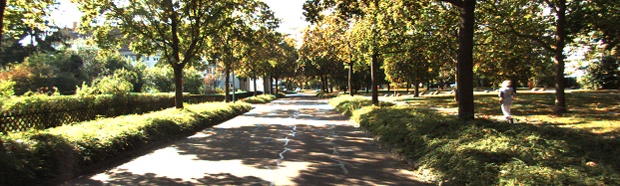}} & 
        \multicolumn{2}{c}{\includegraphics[width=0.22\linewidth]{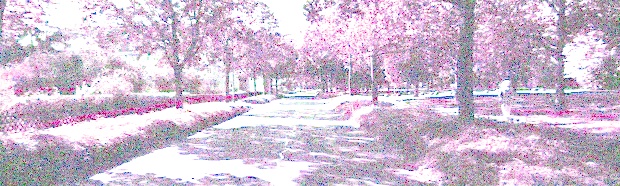}} & 
        \multicolumn{2}{c}{\includegraphics[width=0.22\linewidth]{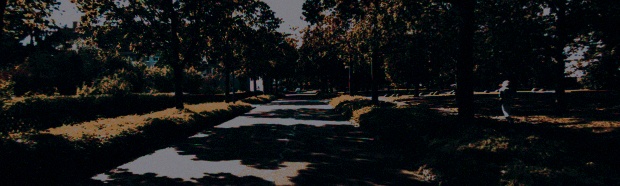}} \\
        \multicolumn{2}{c}{\bf (a) Input} &
        \multicolumn{2}{c}{\bf (b) GT} & 
        \multicolumn{2}{c}{\bf (c) High Intensity} &
        \multicolumn{2}{c}{\bf (d) Low Intensity} \\
        \multicolumn{2}{c}{\includegraphics[width=0.22\linewidth]{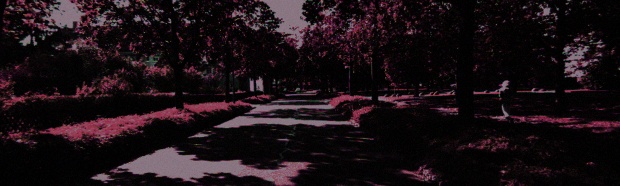}} & 
        \multicolumn{2}{c}{\includegraphics[width=0.22\linewidth]{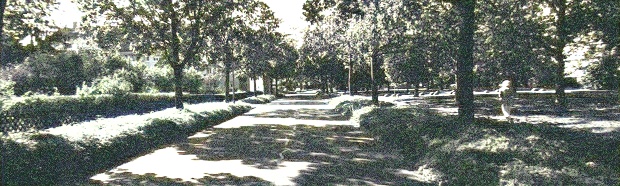}} & 
        \multicolumn{2}{c}{\includegraphics[width=0.22\linewidth]{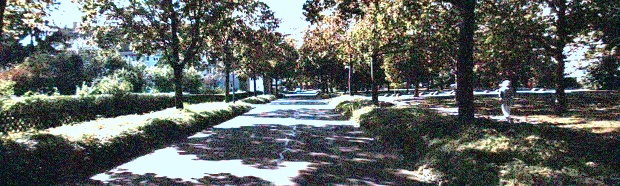}} &
        \multicolumn{2}{c}{\includegraphics[width=0.22\linewidth]{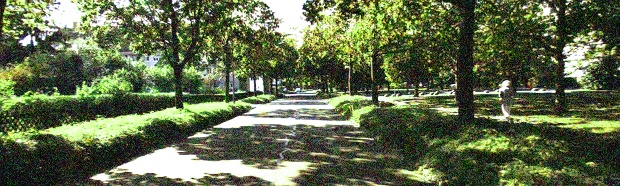}} \\ 
        \multicolumn{2}{c}{\bf (e) Red} &
        \multicolumn{2}{c}{\bf (f) Green} &
        \multicolumn{2}{c}{\bf (g) Blue} &
        \multicolumn{2}{c}{\bf (h) PSNR} \\
        \includegraphics[width=0.105\linewidth]{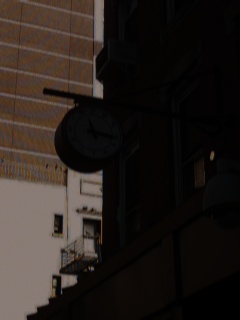} & 
        \includegraphics[width=0.105\linewidth]{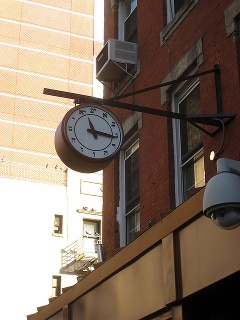} &
        \includegraphics[width=0.105\linewidth]{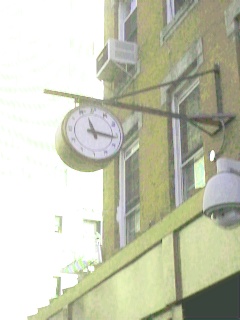} & 
        \includegraphics[width=0.105\linewidth]{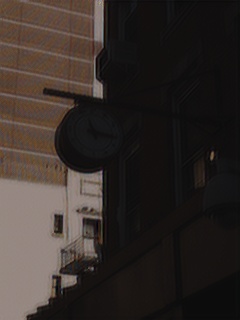} & 

        \includegraphics[width=0.105\linewidth]{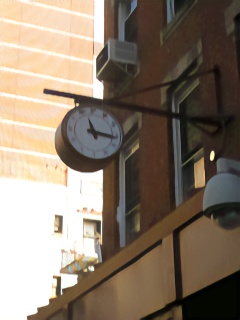} & 
        \includegraphics[width=0.105\linewidth]{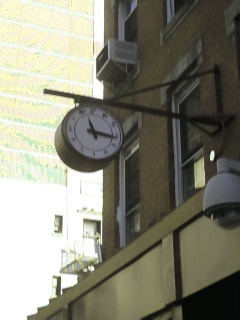} & 
        \includegraphics[width=0.105\linewidth]{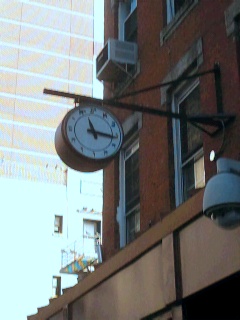} & 
        \includegraphics[width=0.105\linewidth]{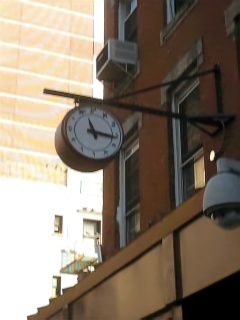} \\
        \textbf{\shortstack{(a) \\Input}} & 
        \textbf{\shortstack{(b) \\GT}} & 
        \textbf{\shortstack{(c) \\High inten}} & \textbf{\shortstack{(d) \\Low inten}} &
        \textbf{\shortstack{(e) \\Red}} & \textbf{\shortstack{(f) \\Green}} & \textbf{\shortstack{(g) \\Blue}} &  \textbf{\shortstack{(h) \\PSNR}}
        \end{tabular}
    \caption{{\bf Qualitative results for RAW-to-RGB processing task with various reward functions.} 
    (a) Input image, (b) GT, (c) High intensity reward, (d) Low intensity reward, (e) Red reward, (f) Green reward, (g) Blue reward, and (h) PSNR reward. The DRL agent tries to maximize the given reward, therefore, the resulting images show diverse results.   }
    \label{fig:raw2rgb_various}
    \vspace{-0.10in}
    \end{figure*}
\end{center}

\subsection{RAW-to-RGB Restoration}
We train the DRL agent with the camera ISP toolbox and PSNR reward function $M_{PSNR}$ to investigate the agent can learn how to restore the RGB image from the given RAW image.
We compare the trained tool selector with the traditional camera ISP pipeline~\cite{abdelhamed2018high} and high-complexity deep neural network, PyNet~\cite{ignatov2020replacing}.
The PyNet is finetuned on the same RAW-to-RGB dataset used for ISP tool selector training during 200 epochs.

The experimental results are quite remarkable in both quantitatively and qualitatively, as shown in~\figref{fig:total RAW2RGB result} and ~\tabref{tab:exp_raw2rgb}.
Even though the tool selector only consists of 2 fully connected layer, the proposed framework effectively generates a clean, sharp, and colorful RGB image from the given image by adaptively selecting proper ISP tools.
As shown in~\figref{fig:teaser}-(c), the tool selector processes an image to maximize the reward function step-by-step according to the current image state.
Furthermore, by adding a color $M_{color}$ or intensity $M_{inten}$ reward, the agent can produces various style of image, as shown in~\figref{fig:raw2rgb_various}. 

\subsection{RAW-to-RGB for Object Detection}
\label{sec:Exp_detection}

We train tool selectors on the MS-COCO~\cite{lin2014microsoft} training set with the object detection reward metrics $M_{PR}$ and $M_{SOPR}$.
After that, the trained tool selectors are evaluated on the MS-COCO~\cite{lin2014microsoft} validation set with GT bounding box labels.
The experimental results are shown in~\figref{fig:exp_det_dep} and~\tabref{tab:exp_obj_det}.
Interestingly, the tool selectors restore a colorful RGB image from the RAW image without the help of PSNR reward metric $M_{PSNR}$.
We believe this is because the detection network (\ie, Mask R-CNN) is trained with colorful RGB images (\ie, MS-COCO), the DRL agent tends to produce an image preferred by the detection network.
As shown in~\tabref{tab:exp_obj_det} and~\figref{fig:exp_det_dep}, the small object aware metric $M_{SOPR}$ shows better detection ability by making DRL-agent produces a small object detail enhanced image.  

\begin{table}[t]
\centering
\caption{\textbf{Quantitative results of RAW-to-RGB processing for object detection task on MS-COCO dataset~\cite{lin2014microsoft}.} 
}
\resizebox{0.90\linewidth}{!}
{
    \def\arraystretch{1.1}
    \begin{tabular}{c|ccc}
    \toprule
    {\bf Method} & {\bf $mAP$} & {\bf $mAP_{50}$} & {\bf $mAP_{75}$} \\
    \midrule
    Default Camera ISP~\cite{abdelhamed2018high} & 24.61 & 36.85 & 26.83 \\
    Ours($M_{PR}$)     & 23.09 & 34.88 & 25.72  \\
    Ours($M_{SOPR}$)   & {\bf 26.40} & {\bf 39.34} & {\bf 29.48}  \\
    \bottomrule
    \end{tabular}
}
\label{tab:exp_obj_det}
\end{table}

\begin{table}[t]
\centering
\caption{\textbf{Quantitative results of RAW-to-RGB processing for single-view Depth Estimation task on KITTI dataset~\cite{geiger2012we}.}
}
\resizebox{0.98\linewidth}{!}
{
    \def\arraystretch{1.2}
    \begin{tabular}{c|cccc|c}
    \hline
    \multirow{2}{*}{Methods} & \multicolumn{4}{c|}{\textbf{Error $\downarrow$}} & 
    \multicolumn{1}{c}{\textbf{Accuracy $\uparrow$}}    \\ \cline{2-6}
     & AbsRel & SqRel & RMS & RMSlog &  $ \delta < 1.25$ \\ 
    \hline \hline
     Camera ISP~\cite{abdelhamed2018high} & 0.131 & 1.009 & 5.323 & 0.210 & 0.839 \\
     Ours($M_{RMSE}$)     & {\bf 0.130} & {\bf 0.978} & {\bf 5.220} & {\bf 0.209} & {\bf 0.839}  \\
     Ours($M_{\delta_1}$) & 0.155 & 1.217 & 5.722 & 0.238 & 0.788 \\
    \hline
    \end{tabular}
}
\label{tab:exp_depth}
\vspace{-0.2in}
\end{table}

\subsection{RAW-to-RGB for Single-view Depth Estimation}
\label{sec:Exp_depth}
\begin{figure}[t]
    \begin{center}
    \begin{tabular}{c@{\hskip 0.002\linewidth}c@{\hskip 0.002\linewidth}c}
     \includegraphics[width=0.30\linewidth]{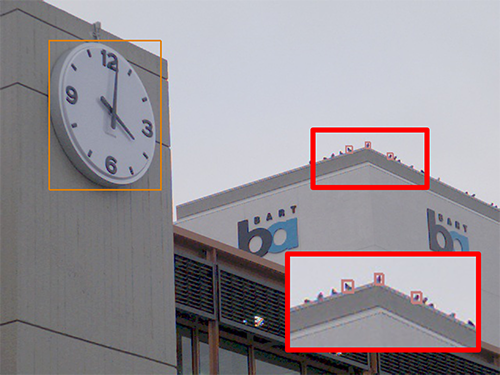} & 
     \includegraphics[width=0.30\linewidth]{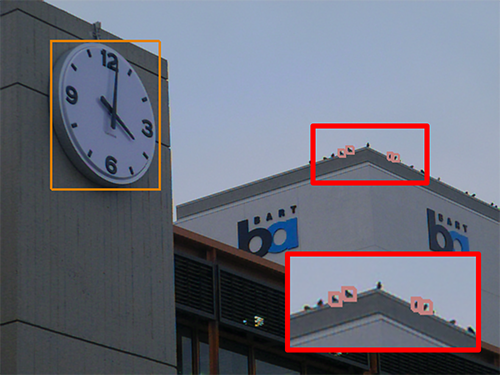} & 
     \includegraphics[width=0.30\linewidth]{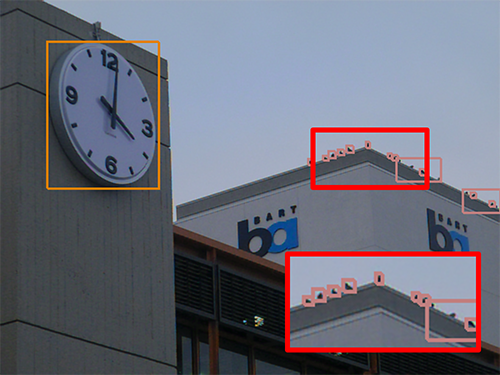} \\ 
     \includegraphics[width=0.30\linewidth]{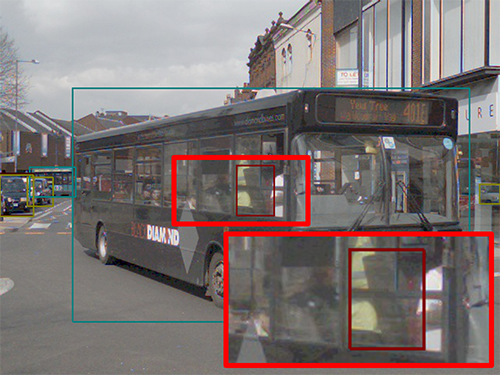} & 
     \includegraphics[width=0.30\linewidth]{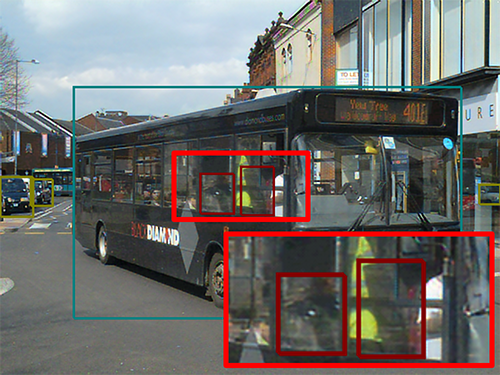} & 
     \includegraphics[width=0.30\linewidth]{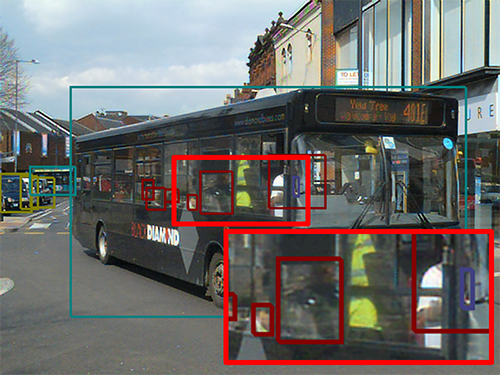} \\ 
     {\footnotesize (a) Default ISP} & {\footnotesize (b) $M_{SOPR}$} & {\footnotesize (c) GT labels}
    \end{tabular}
    \begin{tabular}{c@{\hskip 0.002\linewidth}c@{\hskip 0.002\linewidth}c}
     \includegraphics[width=0.45\linewidth]{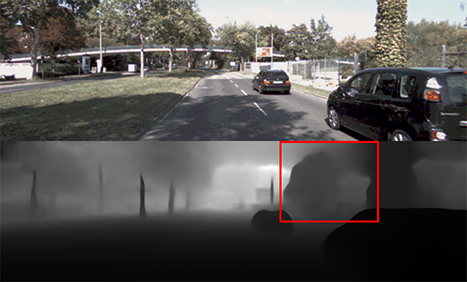} & 
     \includegraphics[width=0.45\linewidth]{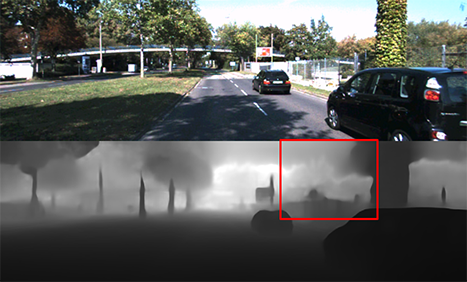} \\ 
     {\footnotesize (d) Default ISP} & {\footnotesize(e) $M_{RMSE}$}
    \end{tabular}
    \caption{\textbf{
    Qualitative results of RAW-to-RGB processing for object detection and depth estimation task.} Our tool selector efficiently processes an input image to maximize given reward with respect to the target task. 
    }
    \label{fig:exp_det_dep}
    \end{center}
    \vspace{-0.2in}
\end{figure}

We train tool selectors on the KITTI~\cite{geiger2012we} training set with the depth reward metrics $M_{RMSE}$ and $M_{\delta_1}$.
After that, the trained tool selectors are evaluated on the KITTI test set with GT depth labels.
The experimental results are shown in~\figref{fig:exp_det_dep} and~\tabref{tab:exp_depth}.
The tool selectors for the depth estimation task also restore a sharp and colorful image according to depth reward function without the help of PSNR reward metric $M_{PSNR}$, same as the object detection task.
The results of~\tabref{tab:exp_depth} show that the metric $M_{RMSE}$ performs better than $M_{\delta_1}$ and default camera ISP.
This indicates that some metrics may not be very effective, even if they are directly supervised by the GT labels, and careful reward function designing is essential for better performance improvements.

\subsection{Ablation Study}
\label{sec:abl}

\begin{table*}[t]
\centering
\caption{\textbf{Ablation Study of the proposed DRL-ISP framework.}
We perform thorough ablation studies on each component of the proposed DRL-ISP framework, such as the architecture of the tool selector, reinforcement learning method, and camera ISP toolbox.
Throughout the ablation study, we validate the effectiveness of the proposed DRL-ISP framework.
}
\centering
\label{tab:abl}
\begin{minipage}{0.48\textwidth}
\centering
    \subfloat[Network Architectures of the tool selector.] 
    {
        \centering
        \resizebox{0.95\linewidth}{!}
        {
            \centering
            \def\arraystretch{1.1}
            \begin{tabular}{c|c|ccc}
            \toprule
            {\bf Type} & {\bf Method} & {\bf PSNR} & {\bf SSIM} & {\bf MSSSIM} \\
            \midrule
            \multirow{9}{*}{\rotatebox{90}{\textbf{Feature Extractor}}} 
            & $N_{\text{DQN}}(\text{learn})$ & 22.40 & 0.6737 & 0.8759 \\
            & $N_{\text{DQN}}(\text{fixed})$ & 22.50 & 0.6818 & 0.8917 \\
            & Intensity $f^{i}$                   & 23.72 & 0.7147 & 0.9057 \\
            & Gradient $f^{g}$                    & 18.87 & 0.6044 & 0.7995 \\
            & Semantic $f^{s}$                    & 24.11 & 0.7147 & 0.9150 \\
            & Inten-Grad $[f^{i},f^{g}]$          & 24.01 & 0.7213 & 0.9130 \\
            & Inten-Sem $[f^{i},f^{s}]$           & 23.50 & 0.7105 & 0.9085 \\
            & Grad-Sem $[f^{g},f^{s}]$            & 23.48 & 0.7004 & 0.8951 \\
            & All $[f^{i},f^{g},f^{s}]$           & {\bf 24.82} & {\bf0.7559} & {\bf0.9375} \\
            \hline
            \multirow{3}{*}{\rotatebox{90}{\textbf{Agent}}} 
            & PolicyNet (4 layer)                & 23.68 & 0.7116 & 0.9006 \\
            & PolicyNet (3 layer)                & 24.46 & 0.7479 & 0.9293 \\
            & PolicyNet (2 layer)                & {\bf24.82} & {\bf0.7559} & {\bf0.9375} \\
            \bottomrule
            \end{tabular}
            \label{tab:abl_arch}
        }
    }
\end{minipage} 
\begin{minipage}{0.48\textwidth} 
\centering
\subfloat[Reinforcement Learning Training Methods.] 
{
    \centering
    \resizebox{0.95\linewidth}{!}
    {
        \centering
        \def\arraystretch{1.15}        
        \begin{tabular}{c|c|ccc}
        \toprule
        {\bf Feature $F$} & {\bf RL Method} & {\bf PSNR} & {\bf SSIM} & {\bf MSSSIM} \\
        \midrule
        $N_{\text{DQN}}(\text{learn})$ &  DQN                    & 6.84 & 0.0379 & 0.1935 \\
        All $[f^{i},f^{g},f^{s}]$ &  DQN         & 10.57 & 0.3852 & 0.6286 \\
        \hline
        $N_{\text{DQN}}(\text{learn})$ &  SAC            & 22.40 & 0.6737 & 0.8759 \\
        All $[f^{i},f^{g},f^{s}]$ & SAC & {\bf 24.82} & {\bf 0.7559} & {\bf 0.9375} \\
        \bottomrule
        \end{tabular}
        \label{tab:abl_rl}
    }
}
\hspace{5mm}
\subfloat[Camera ISP Toolbox.] 
{
    \centering
    \resizebox{0.95\linewidth}{!}
    {
        \centering
        \def\arraystretch{1.0}        
        \begin{tabular}{c|ccc}
        \toprule
        {\bf Tool} & {\bf PSNR} & {\bf SSIM} & {\bf MSSSIM} \\
        \midrule
        $T_{Trad}$                        & 18.70 & 0.5849 & 0.8534 \\
        $T_{CNN}$                         & 24.52 & 0.7439 & 0.9363 \\
        $T_{CNN}(wL_{col})$             & 24.75 & 0.7550 & 0.9323 \\
        $T_{CNN}(wL_{col})$ + $T_{trad}$  & {\bf 24.82} & {\bf 0.7559} & {\bf 0.9375} \\
        \bottomrule
        \end{tabular}
        \label{tab:abl_toolbox}
    }
}
\end{minipage}
\end{table*}

\subsubsection{Network Architecture of Tool Selector}
\label{sec:abl_NetArch}
In this ablation study, we investigate the effectiveness of the proposed tool selector structure.
The proposed tool selector consists of the feature extractor and the DRL agent.
The feature extractor exploits the deterministic feature extraction process from the intensity-, gradient-, and semantic-level.
The experimental results are shown in~\tabref{tab:abl}-(a).
We also compared our feature extraction methods with the DQN network~\cite{mnih2015human} that has learnable parameters and random fixed parameters~\cite{seo2021state}. 
By adding each branch that represents different image properties, the PSNR performance is monotonically increasing.
This phenomenon can be interpreted as each feature vector makes a more explicit relation between the state $s_t$ and action space. 
For example, the gradient feature could be helpful for edge-enhancing action such as sharpening and super-resolution.

The learnable feature extractor $N_{DQN}$(learn) continuously changes the given image's state as the training progress that leads to an unstable training and requires more training steps.
The recent work~\cite{seo2021state} also supports this claim by showing that the random initialized and fixed network $N_{DQN}$(fixed) performs better than the network $N_{DQN}$(learn).
However, we find our explicit state definition $[f^i,f^g,f^s]$ can make more explainable relation than unknown relation $N_{DQN}$(fixed) while showing better experimental results.
We also investigate the effect of a policy network with 2, 3, and 4 fully-connected layers. 
The deeper network does not help increase the performances and hinder the relation learning between the state and actions.

\subsubsection{Reinforcement Learning Method}
\label{sec:abl_RLMethod}
As shown in~\tabref{tab:abl}-(b), even if the same feature extractor $F$ is given, the agent's performance is highly affected by the reinforcement learning methods.
As discussed above, the more efficient exploration-and-exploitation methods~\cite{schaul2015prioritized, lee2020efficient} can greatly increases the performance.

\subsubsection{Camera ISP Toolbox}
\label{sec:abl_Toolbox}
The camera ISP toolbox designing is also an essential factor for the overall performance since the toolbox's status decides the tool selector's maximum and general ability.
As shown in~\tabref{tab:abl}-(c), the traditional tools $T_{Trad}$ can solely enhances the image quality to some extent.
However, the performance improvement is quite limited and is boosted by utilizing CNN-based tools $T_{CNN}$.
The collective tool training $L_{col}$ that enables cascading more than two networks also enhances the performance by alleviating the unseen artifact problem.
The traditional tools $T_{Trad}$ are able to fine-tune the image with controllable parameters and give a supplement effect for CNN-based tools $T_{CNN}(wL_{col})$ that resulting outperformed results. 
The proposed camera ISP toolbox has easily extendable property.
Therefore, we will consider more various ISP tools such as traditional color temperature changing and tone mapping.

\section{Conclusion}
In this paper, we propose a novel multi-objective camera ISP framework that utilizes Deep Reinforcement Learning (DRL) and camera ISP toolbox that consist of a simple network-based tools and conventional tools. 
The proposed ISP toolbox consists of light-weight CNN tools and traditional tools that can represent each block of the default camera ISP pipelines, such as gamma correction, color correction, white balance, sharpening, denoising, and others.
We also provide an efficient DRL network architecture that can extract the various aspects of an image and make rigid mapping relations between image state and a large number of actions.
Our proposed DRL-ISP framework effectively improves the image quality according to various vision tasks such as RAW-to-RGB image restoration, 2D object detection, and monocular depth estimation. 
Furthermore, our framework can generate various styles of images by freely designing reward functions. 
For the future works, we plan to apply our DRL-ISP to our vehicle platforms~\cite{park2019vehicular,rameau2022real} and extend our DRL-ISP to include automatic exposure parameter control~\cite{shin2019camera}.
More visual results are available at~\url{https://sites.google.com/view/drl-isp}.


{\small
\bibliographystyle{IEEEtran}
\bibliography{egbib}
}

\addtolength{\textheight}{-12cm}   



\end{document}